\documentclass[runningheads]{llncs}

\usepackage{times}
\usepackage{xcolor}
\usepackage{soul}
\usepackage[utf8]{inputenc}

\usepackage{comment}
\usepackage{color}
\usepackage{framed}
\usepackage[noadjust]{cite}

\usepackage{epsfig}
\usepackage{graphicx}

\usepackage[ruled,vlined,linesnumbered]{algorithm2e}
\usepackage{url}
\usepackage{multirow}
\usepackage{enumitem}
\usepackage{moresize}
\usepackage{epstopdf}
\usepackage{array}

\usepackage{diagbox}
\usepackage[export]{adjustbox}

\usepackage{tcolorbox}

\usepackage{colortbl}

\usepackage{multirow} 
\usepackage{extarrows}

\usepackage{floatrow}
\newfloatcommand{capbtabbox}{table}[][\FBwidth]

\makeatletter
\newcommand{\thickhline}{%
    \noalign {\ifnum 0=`}\fi \hrule height 1pt
    \futurelet \reserved@a \@xhline
}
\newcolumntype{"}{@{\hskip\tabcolsep\vrule width 1pt\hskip\tabcolsep}}
\makeatother

\usepackage{wrapfig}
\usepackage{subfigure}
\usepackage[hang,flushmargin]{footmisc} 

\usepackage[font=footnotesize,labelfont=bf]{caption}

\usepackage{xspace}

\makeatletter
\DeclareRobustCommand\onedot{\futurelet\@let@token\@onedot}
\def\@onedot{\ifx\@let@token.\else.\null\fi\xspace}

\makeatother

\begin{document}
\begin{sloppypar}
\title{An Orchestrated Empirical Study on Deep Learning Frameworks and Platforms}

\author{Qianyu Guo\inst{1} \and
Xiaofei Xie\inst{2}\and
Lei Ma\inst{3} \and
Qiang Hu\inst{4}\and
Ruitao Feng\inst{2}\and \\
Li Li\inst{5} \and 
Yang Liu\inst{2}\and
Jianjun Zhao\inst{4} \and
Xiaohong Li\inst{1}
}

\institute{Tianjin University, China \and
Nanyang Technological University, Singapore \and Harbin Institute of Technology, China \and
Kyushu University, Japan \and
Monash University, Australia
}

\maketitle

\begin{abstract}
Deep learning (DL) has recently achieved tremendous success in a variety of cutting-edge applications, e.g., image recognition, speech and natural language processing, and autonomous driving. Besides the available big data and hardware evolution, DL frameworks and platforms play a key role to catalyze the research, development, and deployment of DL intelligent solutions. However, the difference in computation paradigm, architecture design and implementation of existing DL frameworks and platforms brings challenges for DL software development, deployment, maintenance and migration. Up to the present, it still lacks a comprehensive study on how current diverse DL frameworks and platforms influence the DL software development process.

In this paper, we initiate the first step towards the investigation on how existing state-of-the-art DL frameworks (i.e., TensorFlow, Theano, and Torch) and platforms (i.e., server/desktop, web, and mobile) support the DL software development activities.
We perform an in-depth and comparative evaluation on metrics such as learning accuracy, DL model size, robustness and performance, on state-of-the-art DL frameworks across platforms using two popular datasets MNIST and CIFAR-10.
Our study reveals that existing DL frameworks still suffer from compatibility issues, which becomes even more severe when it comes to different platforms. 
We pinpoint the current challenges and opportunities towards developing high quality and compatible DL systems. 
To ignite further investigation along this direction to address urgent industrial demands of intelligent solutions, we make all of our assembled feasible toolchain and dataset publicly available.

\end{abstract}

\section{Introduction}
In company with the big data explosion and hardware system evolution over the past decades, deep learning (DL) systems achieved tremendous success and human competitive performance in various cutting-edge applications, such as real-time strategy game~\cite{mnih2015humanlevel, dota2}, image processing~\cite{he2016deep, he2015delving, simonyan2014very}, speech intelligent assistant~(e.g., Siri, Alexa, Cortana)~\cite{wightman2000speech, cheyer2014method}, autonomous vehicle~\cite{arslan2007autonomous, litman2017autonomous}, intelligent manufacturing~\cite{Oztemel2010}, medicine discovery~\cite{shen2017deep} and medical diagnostics~\cite{litjens2016deep}. DL has become the driving force of technology innovation of next generation, penetrating into many domains of industry manufacture process, renovating the industry practice and reshaping almost every aspect of our society and daily life.

A deep neural network (DNN) plays the key role behind the recent success of DL systems. It leverages a data-driven programming paradigm to automatically learn the decision logic from the training data, which is represented in the form of a neural network and connection strengths among neurons. To transfer the learning theory into practice, diverse DL frameworks~(e.g., TensorFlow~\cite{tensorflow}, PyTorch~\cite{paszke2017automatic}, and Theano~\cite{team2016theano}) are developed towards realizing urgent industrial demands of intelligent solutions. Although most of existing DL frameworks share either static or dynamic computational paradigms~\cite{DBLP:journals/corr/AbadiABBCCCDDDG16,DBLP:journals/corr/LooksHHN17}, the detailed architecture design and implementation across frameworks are quite different. In other words, even the same DNN architecture design with exactly the same runtime configuration might result in different decision logic, when implemented under different DL frameworks. Until now, unfortunately, it still lacks a comparative study on what different impacts various frameworks exert on the DL software development process and activities.

With the rapid market trend in developing and deploying the AI solutions to platforms such as mobile devices, web services, and edge computing devices, it further poses challenges when DL systems are migrated, customized and deployed across platforms, considering the diverse requirements and limitations of each platform. For example, while a computational intensive DL system could execute efficiently on PC or server with the GPU support, it might still be an inappropriate scheme when deployed on mobile devices where limited computing capacity and battery volume are available. 
Therefore, some DL frameworks are specifically designed for mobile platforms, such as the recently released DL runtime framework \texttt{TensorFlow Lite}~\cite{Tensorflowlite} for Android and \texttt{Core ML}~\cite{coreml} for iOS. Similarly, variants of DL frameworks~(e.g., \texttt{TensorFlow.js}~\cite{tensorflowjs}) catering for web application are also proposed.

For traditional softwares, we have gone through compatibility and migration issues across platforms for many years, with extensive studies performed regarding issues across programming languages~\cite{Terekhov:2000:RLC:624640.626213}, platforms, and hardware systems~\cite{Faust03softwareproduct, Linden:2007:SPL:1296141}. 
When it comes to DL software, similar compatibility and migration issues, regarding heterogeneous DL frameworks, platforms and hardware systems, still persist or even potentially more severe due to the immature development and deployment DL framework supports. It is nowadays not clear how well do different platforms support DL applications in terms of accuracy, computational efficiency, energy consumption, robustness, etc.

However, due to the significant difference of DL software from traditional software in terms of programming paradigm, decision logic and software artifact representation, the accumulated experience and well established industrial best practice on traditional software could not be directly applied to DL application scenarios~\cite{2018arXiv181004538M, ma2018deepgauge, ma2018deepmutation, 2018arXiv180605859B}.
Although the DL development framework and platform supports are experiencing rapid evolution driven by the urgent market demands, it is still far from a mature state for large-scale application. 
There even lacks a comparative study on how different DL frameworks and platforms impact the development, deployment, migration, as well as hardware adaption of DL applications. It also lacks an assembled and manually validated toolchain to allow a systematic study of DL software development on different frameworks and platforms.
To fill this gap, in this paper, we perform an orchestrated empirical study on current state-of-the-art deep learning frameworks and platforms, and provide a toolchain framework to make a large-scale comparative study feasible. In particular, we aim to investigate the following research questions:

\begin{itemize}
    \item \emph{\textbf{RQ1}}: Given the same DNN design and runtime training configuration, does it exhibit different runtime behavior and training performance when implemented under different DL frameworks?
    
    \item \emph{\textbf{RQ2}}: When DL models with the same design and runtime configuration are trained under different DL frameworks, what on earth are their prediction performance in terms of accuracy and time efficiency under different DL frameworks?
    
    \item \emph{\textbf{RQ3}}: Does the robustness of DL models trained from different frameworks appear to be the same?

    \item \emph{\textbf{RQ4}}: Given the same model, what are their prediction performance in terms of accuracy and time efficiency under different DL platforms? How do platform customization and optimization during cross-platform migration and deployment influence the DL prediction performance?
\end{itemize}

Through answering these questions, we aim to characterize the potential impact of current DL frameworks and platforms on DL software development and maintenance process, and provide useful findings to DL software developers, researchers and practitioners. In summary, the major contributions of this paper are as follows:

\begin{itemize}
    \item To the best of our knowledge, this paper is the first large-scale comparative study on how current DL frameworks and platforms influence the development, deployment, and migration of DL applications. We highlight current best practices, challenges, and point out opportunities. 
    
    \item Our study reveals that different DL frameworks indeed exhibit different training performance, resulting in different DL models with different size, prediction performance, etc. Our further investigation finds that the robustness of obtained DL models from different DL frameworks are also different, although they share the same design and runtime training configuration. Moreover, platform customization and optimization would introduce potential defects and issues, which need careful assessment before deployment.
    
    \item We make all our dataset and assembled toolchain publicly available, to facilitate further study towards more systematic investigation. We hope our work draws the attention of research community on the currently important development and deployment issues of DL systems, altogether to address the urgent industrial needs.
\end{itemize}
\section{Background}
\subsection{Deep Neural Network}
Inspired by how the biological nervous systems function to process information as a decision system, DNN is designed to share the similar structure as a neural network. In particular, a DNN usually consists of multiple layers of computing units (i.e., neurons): a layer of input neurons (i.e., input layer), one or multiple layers of hidden neurons (i.e., hidden layers) and a layer of output neurons (i.e., output layer), with its neurons connected between layers to propagate the information for further processing.

Different from traditional software whose decision logic is often manually programmed, DL defines a data-driven programming paradigm that automatically learns its decision logic through a large set of training data, and encode them into the (1) network architecture and (2) connection weights between neurons and layers. The major effort for a DL developer is to prepare the training data, design the DNN architecture and specify the runtime training configuration as a training program, after which the decision logic is automatically learnt during the training process. The training process is often a highly computational intensive and time consuming optimization procedure. The recent DL software stack support of training on GPU and Spark based distributed system~\cite{2018arXiv181004538M} makes deep learning development available to everyone, which could only be achieved by specialized high performance computing machine several years ago.

\subsection{Deep Learning Frameworks}\label{sec2:dlframe}
Deep learning framework plays an important role to bridge the DL theory to the realization of DL software, which provides high level APIs to support the DNN design and runtime training configuration.
Over the past decade, many DL frameworks are proposed and developed from academic (e.g. Berkeley and NYU) and industry (e.g. Microsoft, Google, Facebook and Nvidia), such as Tensorflow~\cite{tensorflow}, CNTK~\cite{seide2016cntk}, Torch~\cite{collobert2011torch7}, Theano~\cite{team2016theano}, Caffe~\cite{jia2014caffe}, Chainer~\cite{chainer}, Deeplearning4j~\cite{deeplearning4j}, etc. 
Almost all existing state-of-the-art DL frameworks take the hardware acceleration into account during design stage, in order to boost the training and validation performance. However, these DL frameworks follow various architecture designs and computational paradigms, with the implementation via different programming languages~(e.g., Python, C/C++, Java and Lua). Furthermore, even the same DL training algorithm might be implemented distinctly under the adopted computational paradigm.
Therefore, it is highly possible that the same training data, DL model design and training configuration might still exhibit different training performance, and result in DL models with diverse decision logics. The developed DL models under different frameworks also confront with compatibility issues, causing a DL model hard to migrate from one DL framework to another. 
Although some recent work~\cite{MMdnn, onnx} initiated to concern the DL framework compatibility problem, towards enabling the interoperability of DL models among frameworks, it is still in infancy with many conversion limitations and issues.
\footnote{\scriptsize{For example, MMdnn~\cite{MMdnn} DL model exchange library oftentimes could not preserve the model structure during conversion from Kears to Pytorch framework; and some parameters might also not be exactly preserved therein, causing the conversion failure or obtained DL model with low performance.}} 

\subsection{Deep Learning Platforms Customization and Optimization}
While most existing DL frameworks~(e.g., see studied frameworks in Section~\ref{sec2:dlframe}) support general desktop DL software development and execution, the recently rapid development of system-on-chip (SoC) acceleration~(e.g., Qualcomm Snapdragon, Kirin 970, Samsung Exynos9) for AI applications provides the hardware support and foundation for universal deployment~\cite{2018arXiv181001109I} across platforms, especially on mobile device, edge computing device and so forth.
However, due to the computation power, memory size, and energy limitation of mobile and edge device liked platform, the DL frameworks for PC platform could not be directly transplanted. Recently, some lightweight solutions are proposed for mobile platforms such as CoreML~\cite{coreml}, TensorFlow Lite~\cite{Tensorflowlite}, Caffe2 Mobile~\cite{caffemobile} and Torch Android~\cite{torchandroid}. Likewise, a solution is also proposed for deploying DL models in the web environment~(e.g., TensorFlow.js). 
Even though, the current best industrial practices still perform development of DL software in the PC/server environment, after which the obtained DL models are customized to deploy on specific target platform.
For example, it is a common practice that a DL model needs to go through the quantization and compression phase before deploying on mobile devices, considering the limited resource for memory and energy on target platform. There exists a nonnegligible challenge that such customization and optimization might reduce the prediction accuracy, and even bring in unexpected runtime behaviors, compared to the original model. Because of that, we also study what impact the platform customization could exert on the DL runtime performance. 

\section{EMPIRICAL STUDY METHODOLOGY} \label{sec:methodology}
\begin{figure*}
    \centering
	\includegraphics[width=1.0\textwidth]{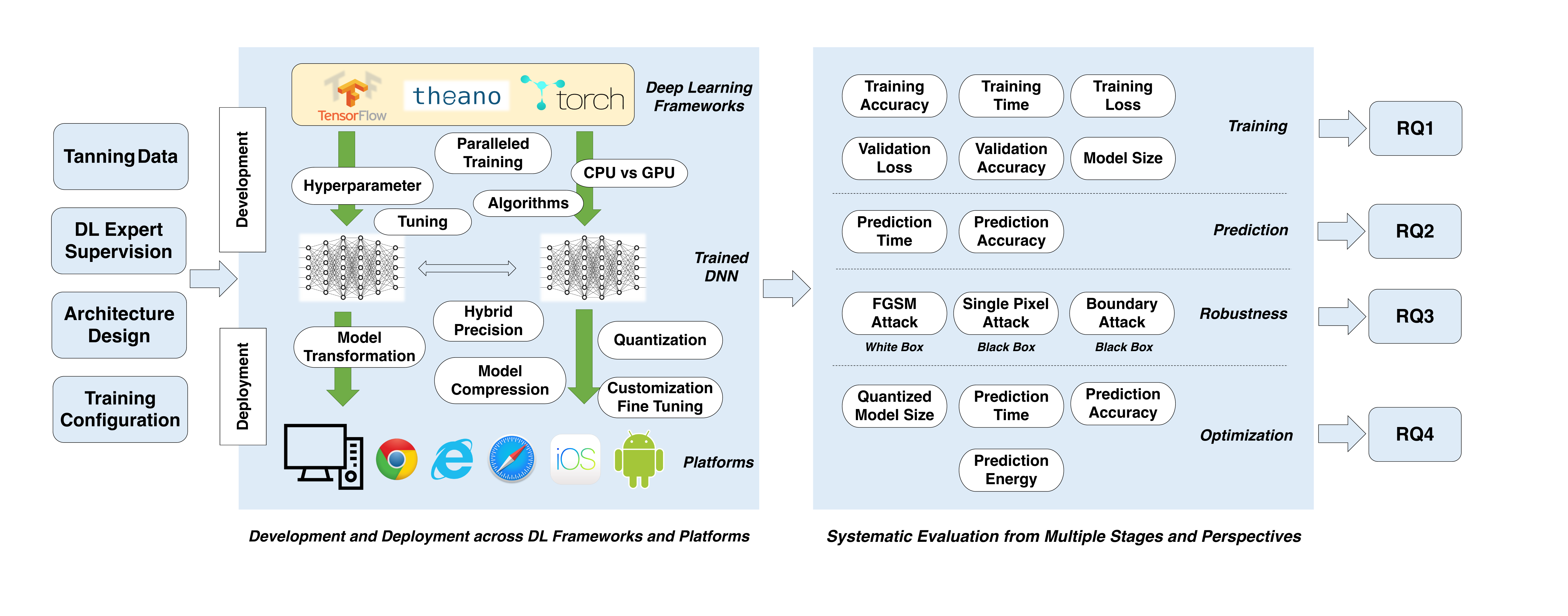}
	\caption{The overview of the empirical study design, workflow and evaluation indicators among different stages of development and deployment of DL software. }
	\label{fig:overview}
\end{figure*}

Figure~\ref{fig:overview} summarizes the overall workflow of our empirical study. There are several key phases, including two preparation steps (1) select subject datasets as well as the DL models (Section~\ref{sec:modelselect}), (2) select the state-of-the-art DL frameworks and execution platforms for investigation (Section~\ref{sec:frameworkselect}); (3) based on this, we perform in-depth systematic evaluation on different phases of DL development process~(i.e., training, prediction execution, robustness evaluation, platform optimization and customization, see Section~\ref{sec:metrics}), towards answering the four concerned research questions.

\subsection{Datasets and Models} \label{sec:modelselect}
The dataset selection is always an important factor concerning for an empirical study. In this work, we select two publicly available datasets (i.e., MNIST~\cite{mnist} and CIFAR-10~\cite{cifar}) as the evaluation subject datasets for training and prediction, both of which are widely used in deep learning and software engineering research community. For each dataset, we follow the best DL practice and choose the DNN models~\cite{lenet-fam,2018arXiv180102610X,carlini2017towards} that are able to achieve competitive performance in terms of training and validation accuracy.

\noindent\textbf{MNIST} is a collection of single-channelled image data of size $28 \times 28 \times 1$ for 10 hand-written digit recognition. MNIST contains $70,000$ data in total, consisting of $60,000$ training data and $10,000$ test data. Each MNIST image is $28 \times 28 \times 1$ in size. On MNIST, we select two well-known models from the LeNet family (i.e., LeNet-1 and LeNet-5~\cite{lenet-fam}) for comparative study. Table~\ref{tab:subject_sum} summarizes the complexity of each studied DL models in terms of the number of neurons, layers, and the trainable parameters.

\noindent\textbf{CIFAR-10} is a collection of images for general-purpose image classification~(e.g., airplane, automobile, bird, cat), including $60,000$ colour images in total with $50,000$ training images and $10,000$ test images, respectively. Each CIFAR-10 image is three-channel of size $32 \times 32 \times 3$, which is about 4 times in dimensionality of MNIST image. Therefore, the DL models are often complex in depth and network architecture to obtain competitive performance. On CIFAR-10, we select two popular DNN models (i.e., ResNet-20~\cite{resnet} and VGG-16~\cite{vgg}) for inspection, both of which could achieve competitive prediction performance.

\begin{table}[]
\centering
\caption{Summary of Subject Dataset and DL Models}
\label{tab:subject_sum}
\begin{tabular}{|c|c|c|c|c|c|}
\hline
Dataset & \begin{tabular}[c]{@{}c@{}}Dataset\\ Description\end{tabular} & DNN Model & \#Neuron & \#Layer & \begin{tabular}[c]{@{}c@{}}Train. Para.\end{tabular} \\ \hline\hline
\multirow{2}{*}{MNIST} & \multirow{2}{*}{Digit recog} & LeNet-1 & 52 & 7 & \multirow{2}{*}{\begin{tabular}[c]{@{}c@{}}3,246\\
44,426 \end{tabular}} \\ \cline{3-6}
 &  & LeNet-5 & 268 & 9 &  \\ \hline
\multirow{2}{*}{CIFAR-10} & \multirow{2}{*}{\begin{tabular}[c]{@{}c@{}}General image\\ with 10 classes\end{tabular}} & ResNet-20 & 2,570 & 70 & \multirow{2}{*}{\begin{tabular}[c]{@{}c@{}} 273,066\\ 33,663,070 \end{tabular}} \\ \cline{3-6}
 &  & VGG-16 & 12,426 & 17 &  \\ \hline
\end{tabular}%
\end{table}

\vspace{-5ex}
\subsection{Deep Learning Frameworks and Platforms} \label{sec:frameworkselect}
For DL frameworks, we select three popular and widely used frameworks: TensorFlow~\cite{tensorflow}, Theano~\cite{team2016theano}, and Torch~\cite{collobert2011torch7} as our subject DL frameworks for investigation, where the former two adopt the static computational graph paradigm, while Torch follows a dynamic computation paradigm.

\noindent\textbf{TensorFlow} is an open source framework for high performance numerical computation, which is originally developed by researchers and engineers from the Google Brain team within Google’s AI organization. TensorFlow computations are expressed as static, symbolic stateful dataflow graphs, defined by a set of tensors that themselves are multidimensional data arrays.

\noindent\textbf{Theano} is a Python DL framework that was originally developed as a symbolic math processor at the University of Montreal. It supports defining, optimizing and evaluating mathematical expressions involving high dimensional tensors. Hence, Theano is widely adopted by neural network and machine learning researchers for computing the gradients of an error function with respect to the weights of a network. 

\noindent\textbf{Torch} is an open source machine learning library, which was originally developed at NYU and supported by Facebook at this moment. It is implemented based on Lua programming language and provides a wide range of algorithms for deep learning. Different from the static "define-and-Run" paradigm in TensorFlow and Theano, Torch leverages dynamic "define-by-Run" computational paradigm so as to provide more flexibility, modularity, and readability.

To study the platform influence on DL software development and deployment, we select to cover the three major platforms, where an urgent demand on DL software solutions exists.
\begin{itemize}
	\item The general-purpose {\it desktop/server platform} (also known as {\it PC}), where most DL software are trained.
	\item The {\it mobile platform}. The support of DL software execution on mobile devices todate become available with the advent of lightweight DL frameworks such as \texttt{TensorFlow Lite} for Android, and \texttt{Core ML} for iOS.
	\item The {\it web platform} which is designed to run on the browser environment. We find a feasible framework support, i.e. \texttt{TensorFlow.js}, which enables the DL software execution on dynamic web applications.
\end{itemize}

\subsection{Large-Scale Comparative Evaluation} \label{sec:metrics}
With the aforementioned datasets, we perform a large scaled comparative study on three DL frameworks and three platforms with a total of $14$ evaluation metric configurations from four different perspectives. The adopted metrics are used to evaluate the performance on DL training, prediction, robustness, DL model customization and optimization~(see Figure~\ref{fig:overview}). The detailed configuration of studied metrics will be discussed in Section~\ref{sec:results}.

\begin{itemize}
\item \textbf{Training}. To evaluate how various DL frameworks behave in its training period, we setup a controlled training environment. In particular, we prepare the same DNN design and runtime training configuration, and perform large scale trainings on different DL frameworks for comparative evaluation by observing the following metrics: 1) training/validation accuracy that measures the prediction accuracy on training/validation data during the training phase, 2) training/validation loss that represents the optimization error on the training/validation data in training, 3) training time that constructs and updates the DNN decision logic under different learning iteration~(i.e., epochs) and 4) model size that indicates the storage and runtime memory consumption of trained DL models on different frameworks.

\item \textbf{Prediction}. After training process, we compare the prediction performance of the DL models with two metrics: 1) prediction time that measures the time spent on each prediction execution, and 2) the prediction accuracy on all of the accompanied test data. These two metrics enable us to estimate the performance of a deployed DL model under different DL frameworks.

\item \textbf{Robustness}. Robustness is an important factor indicating the quality of the trained DL models~\cite{ma2018combinatorial}. Given an input $\textbf{x}$ to a DNN, the DL robustness property is concerned with whether there exists another input $\textbf{x'}$ close enough to $\textbf{x}$, with respect to some distance metrics~(e.g., $L_{0}$\text{-}norm, $L_{\infty}$\text{-}norm), that $\textbf{x}$ and $\textbf{x'}$ are classified to different classes by the DNN. Such an input $\textbf{x'}$, once exists, is called an adversarial example of $\textbf{x}$ and the DNN is not robust at $\textbf{x}$. Let $\mathcal{C}(\textbf{x})$ denote the category to which $\textbf{x}$ is classified by DNNs. Formally, a DNN is $d$-robust at an input $\textbf{x}$ w.r.t a distance parameter $d$ iff we have the following relation~\cite{GCDKM17}:
$$\forall \textbf{x'}: ||\textbf{x'}-\textbf{x}||\leq d \Rightarrow \mathcal{C}(\textbf{x})=\mathcal{C}(\textbf{x'})$$  

We evaluate the robustness of a trained model on how resilient it is against adversarial attacks. Specially, we apply a state-of-the-art white-box attack and two black-box adversarial attacks on the trained model, to investigate the potential influence from different DL frameworks and platforms on model robustness.

\item \textbf{Optimization}. 
Current deep learning softwares are mainly used to solve general purpose classification and prediction problems. In practice, the input dimensions are often quite large. For example, even the the relatively small data MNIST has 784 dimensions for the input layer. It is often the case that a DL model with a complex structure and large number of trainable parameters could achieve competitive results on more general purpose dataset~(e.g., CIFAR-10 that has 3072 dimensions). However, such large DL models would take large power and memory consumption meanwhile. To deploy such models on resource limited mobile devices, the usual practice is to go through a quantization or optimization procedure to ensure smooth migration to the target platform. To be specific, quantization is a technique pruning the DL model to reduce model size, so as to increase its inference efficiency with lower power consumption. We simulate the current best practice for mobile DL model deployment, and study how such a quantization influences the model migration.
\end{itemize}
\section{Empirical Study Results} \label{sec:results}
In this section, we conduct numerous comparative experiments to answer the research questions mentioned above. To support such a large-scale study, we run all the desktop application experiments on a high performance computer cluster. Each cluster node runs a GNU/Linux system with Linux kernel 4.4.0 on a 28-core 2.3GHz Intel Xeon 64-bit CPU with 196 GB  RAM equipped with a NVIDIA Tesla V100 16G-GPU. Web application experiments are conducted on a laptop with 32-bit Google Chrome 70.0.3538.67. The mobile application experiments are conducted on Android devices with Huawei Nexus 6P, Motorola Nexus 6, and an iOS device with IPhone 6S. 

\subsection{RQ1: Training Performance} \label{sec:training-performance}
This section investigates the performance of training process hosted on different deep learning frameworks. To answer the aforementioned question, four models (LeNet-1 and LeNet-5 for MNIST, ResNet-20 and VGG-16 for CIFAR-10) are trained on each framework. For each model, the network structure and training parameters are guaranteed as the same under each framework.

We compare the training performance of different frameworks on different settings. Specifically, each model will be trained under different training batch sizes (i.e., 32, 64 and 128), and different processing units (i.e., CPU and GPU). It is also notable that each model, with one set of setting parameters, is repeatedly trained 5 times, and the average of metrics are used for comparison.

\begin{table*}[!t]
\centering
\caption{Average model size and training time with different settings}
\label{tab:traininigtime}
\small
\resizebox{1.0\textwidth}{!}{%
\begin{tabular}{|c|c|c|c|c|c|c|c|c|c|c|c|c|}
\hline
\multirow{3}{*}{Model} & \multicolumn{4}{c|}{TensorFlow} & \multicolumn{4}{c|}{Theano} & \multicolumn{4}{c|}{Torch}                                                                                 \\ \cline{2-13} 
& \multirow{2}{*}{\begin{tabular}[c]{@{}c@{}}Model \\ Size\end{tabular}} & \multicolumn{3}{c|}{ Time Per Epoch (s)} & \multirow{2}{*}{\begin{tabular}[c]{@{}c@{}}Model\\ Size\end{tabular}} & \multicolumn{3}{c|}{Time Per Epoch (s)} & \multirow{2}{*}{\begin{tabular}[c]{@{}c@{}}Model\\ Size\end{tabular}} & \multicolumn{3}{c|}{Time Per Epoch (s)} \\ \cline{3-5} \cline{7-9} \cline{11-13} 
&     & B-32    & B-64  & B-128       &  &B-32         & B-64       & B-128  & &B-32         & B-64        & B-128       \\ \hline\hline
LeNet-1 & 16KB  & 8.50  &   4.59    &   2.29    &   65KB    &   14.84   & 11.83      & 9.76     & 14KB  & 9.09       & 6.57      & 6.81      \\ \hline
LeNet-5 &   178KB   & 11.80     & 5.13  & 2.74  & 556KB     &16.13     & 6.85   &   5.71    & 176KB & 9.14    & 7.23  & 6.99      \\ \hline
ResNet-20   &   1.1MB   & 33.95     & 19.38     & 14.35     & 2.4MB     &   574.72  &  573.28   &   575.21  & 1.1MB  & 30.87      & 16.11     & 11.65     \\ \hline
VGG-16  &   129MB   & 63.87      & 38.98     & 26.17    &   258MB   &   249.85  &  262.50   &    249.69 & 129MB &   49.12   & 34.02     & 28.18     \\ \hline
\end{tabular}%
}
\end{table*}

\begin{figure}[t]
	\centering
	\subfigure[LeNet-5]{
		\label{fig:lenet5-trloss}
		\includegraphics[width=0.48\textwidth]{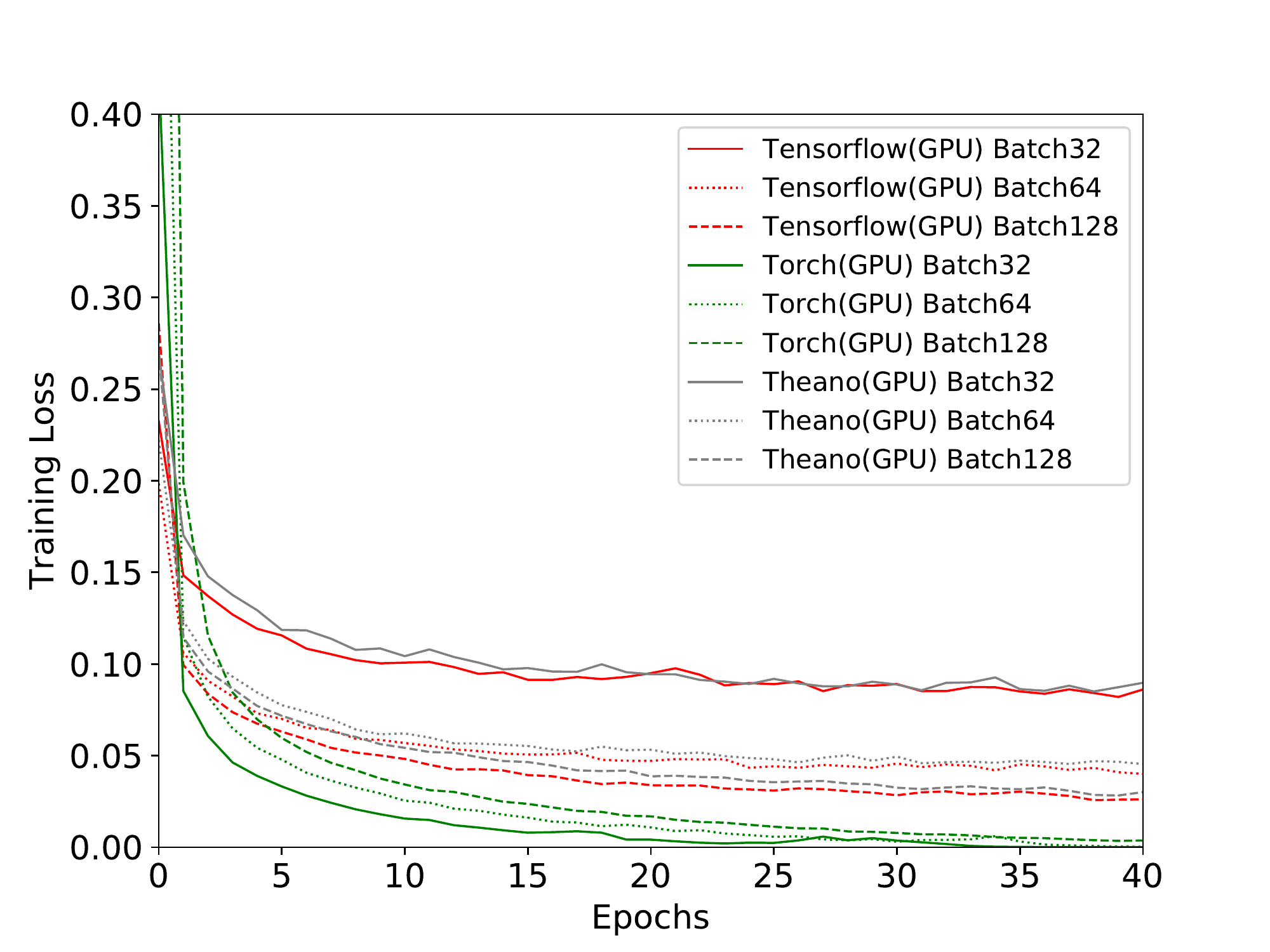}
	}
	\subfigure[ResNet-20]{
		\label{fig:resnet20-trloss}
		\includegraphics[width=0.48\textwidth]{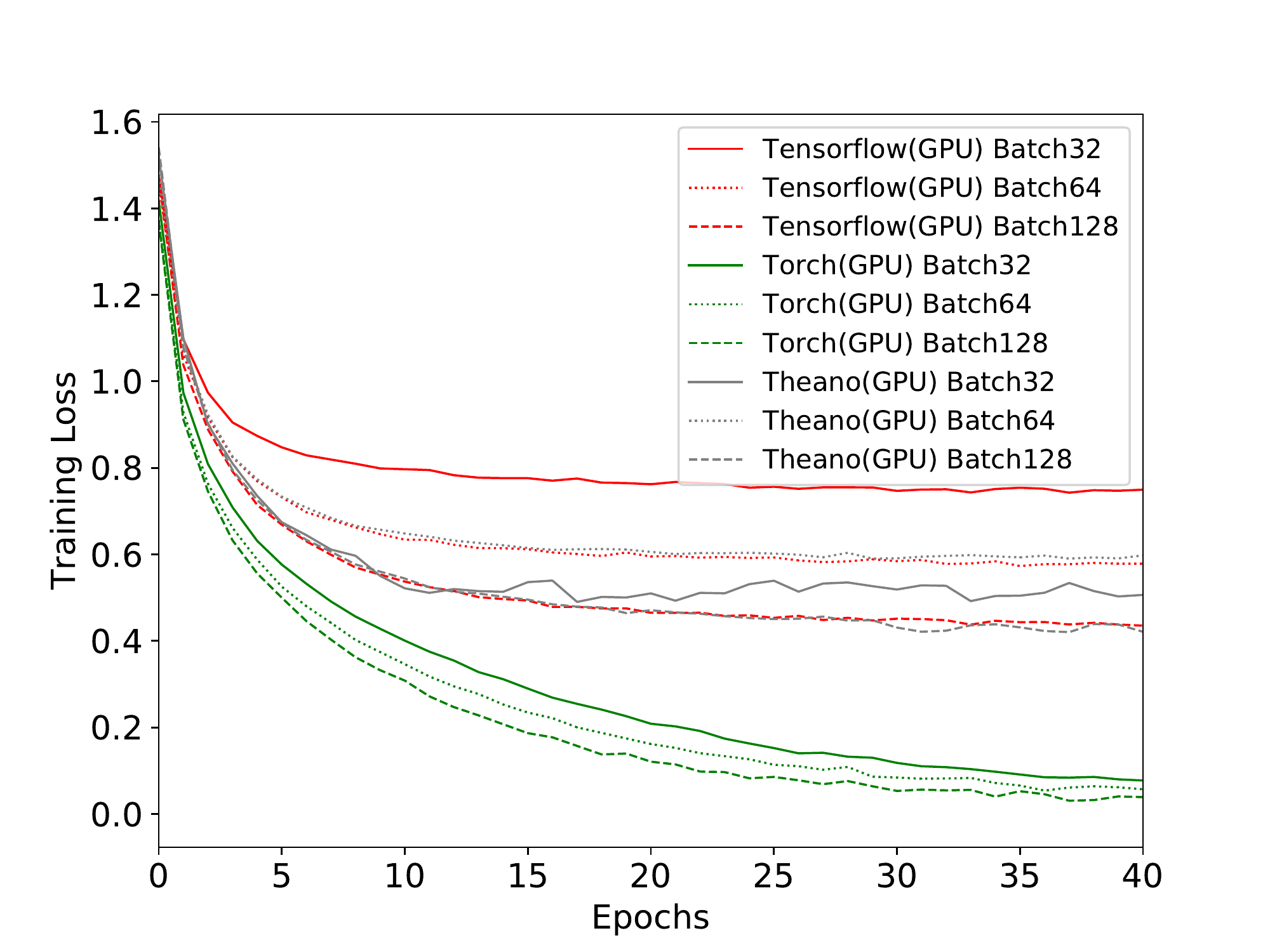}
	}
	\caption{Training loss of LeNet-5 and ResNet-20 with different training batch sizes and frameworks running on GPU}
	\label{fig:batchsizelosseval}
\end{figure}

\noindent\textbf{Training Loss}.
Figure~\ref{fig:batchsizelosseval} shows the training loss plots during training LeNet-5 and ResNet-20 on GPU with different training batch sizes and frameworks. From the training loss curves, we first study the influence of different batch sizes for each framework. In details, training LeNet-5 on TensorFlow and Theano, with bigger batch size, the training loss will be smaller. Conversely, for Torch, with smaller batch size, the training loss will be smaller.In ResNet-20, for all frameworks, training loss will be smaller when the batch size is larger.
In addition, by comparing the training loss among different frameworks, we can observe that training loss of Torch will be smaller than TensorFlow and Theano.

\begin{figure}[t]
	\centering
	\subfigure[LeNet-5]{
		\label{fig:lenet5-tracc}
		\includegraphics[width=0.48\textwidth]{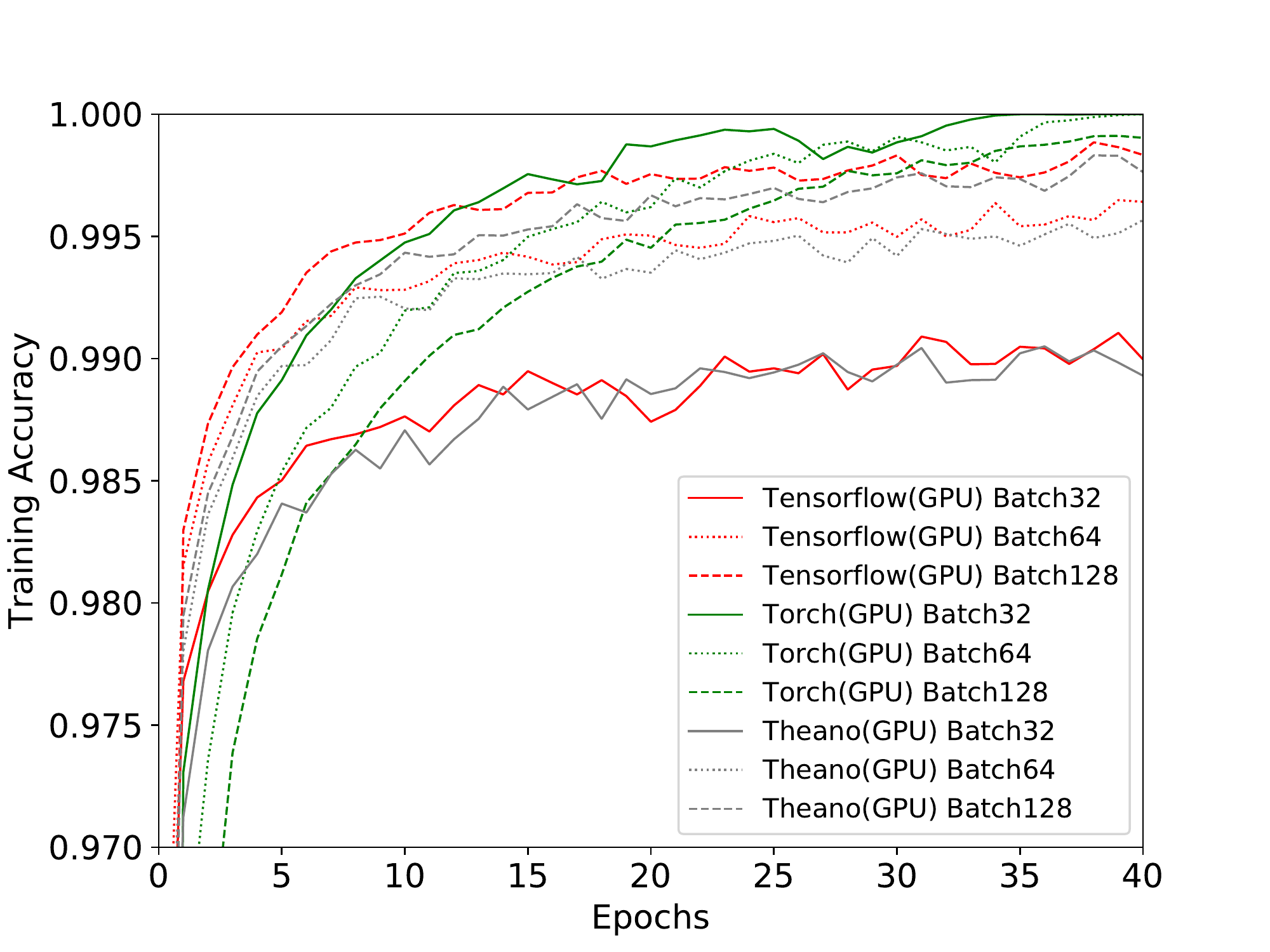}
	}
	\subfigure[ResNet-20]{
		\label{fig:resnet20-tracc}
		\includegraphics[width=0.48\textwidth]{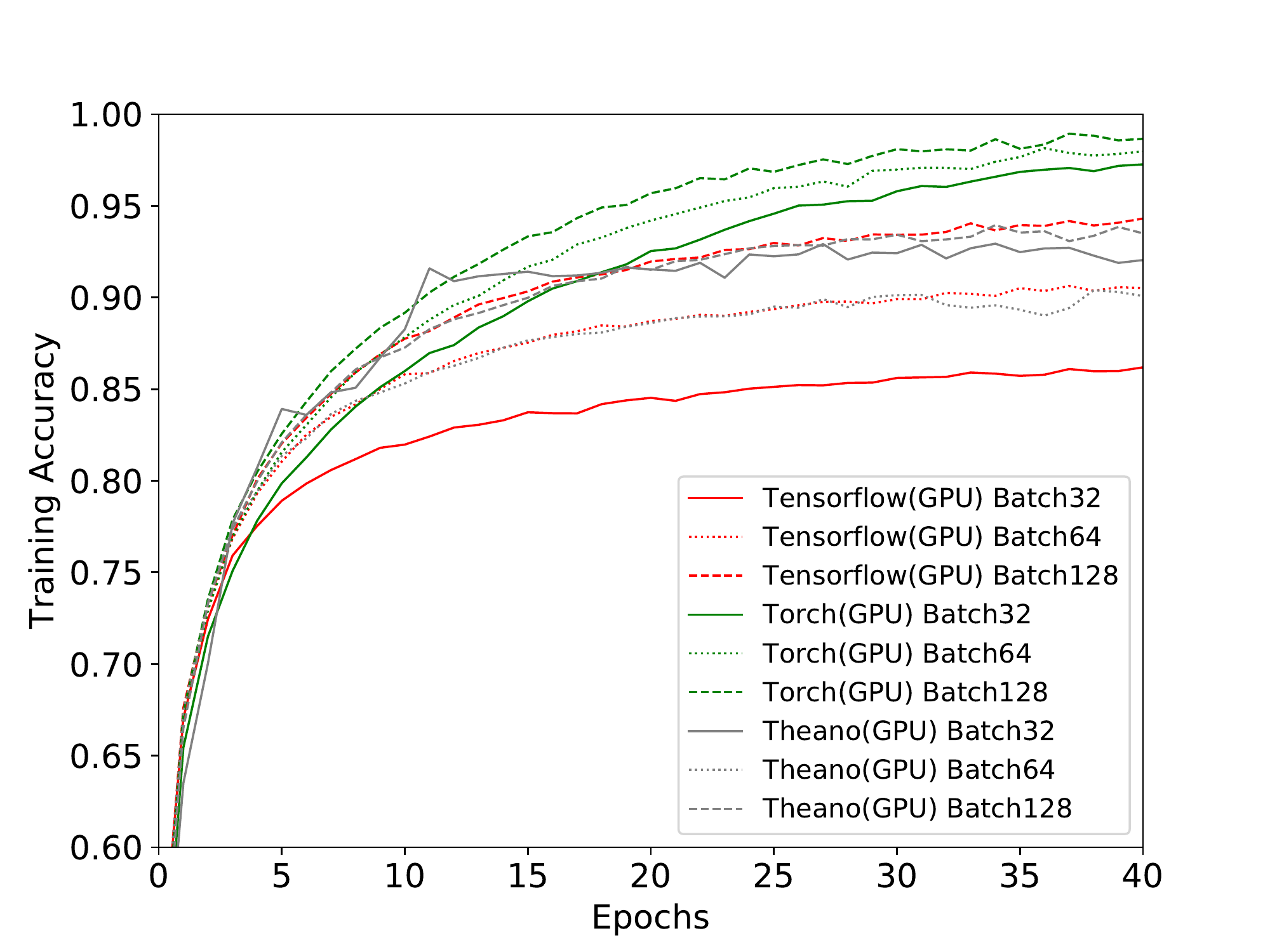}
	}
	\caption{Training Accuracy of LeNet-5 and ResNet-20 with different training batch sizes and frameworks running on GPU}
	\label{fig:batchsizeacceval}
\end{figure}

\noindent\textbf{Training Accuracy}.
Figure~\ref{fig:batchsizeacceval} shows the training accuracy plots that correspond to the training loss plots in Figure~\ref{fig:batchsizelosseval}. When training loss is smaller, the training accuracy will be higher. For TensorFlow and Theano, the training accuracy becomes higher in LeNet-5 and ResNet-20, when the batch size is bigger. For Torch, in LeNet-5, smaller batch size makes the training accuracy higher while the situation is just contrary in ResNet-20. Overall, Torch achieves higher training accuracy than TensorFlow and Theano in LeNet-5 and ResNet-20. From the curve in Figure~\ref{fig:lenet5-trloss} and Figure~\ref{fig:lenet5-tracc}, we also observe that Torch with batch size 32 has the minimum training loss when the epoch is 5, but TensorFlow with batch size 128 has the highest training accuracy in epoch 5, which indicates that training loss in different frameworks seems to follow different mechanisms.

\begin{figure}
    \centering
    \includegraphics[width=0.5\textwidth]{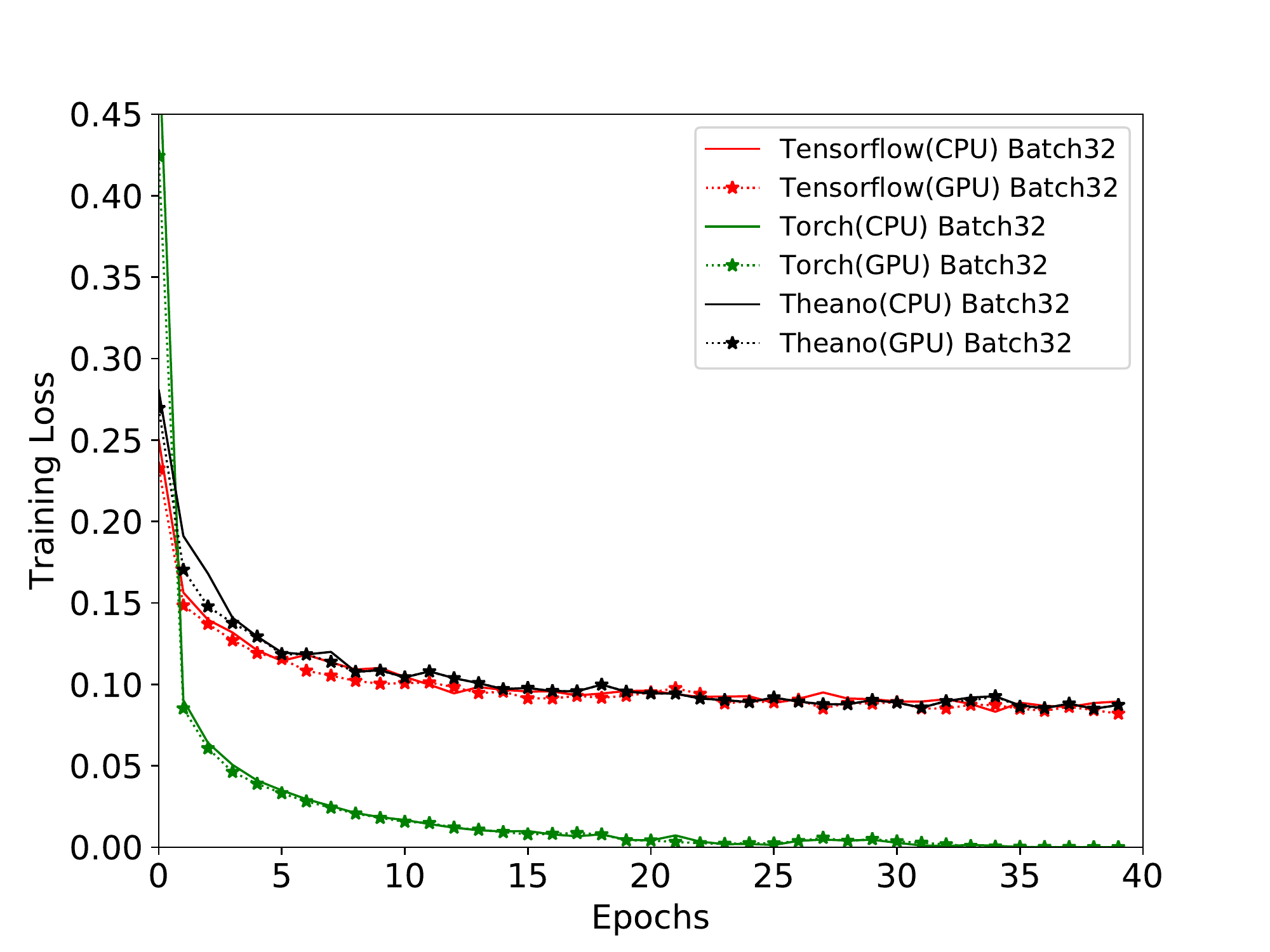}
    \caption{Training loss of LeNet-5 with batch size 32 on CPU and GPU}
    \label{fig:cpugpu}
\end{figure}

\noindent\textbf{CPU vs GPU}.
We compare the training loss and accuracy of other models on CPU and GPU, the results are indeed similar between CPU and GPU. Figure~\ref{fig:cpugpu} shows the training loss plots for LeNet-5 with batch size 32 on CPU and GPU, respectively. We can observe that the training loss are very close on CPU and GPU for each framework. Details of other models can be referred to on our website.

\begin{figure}[t]
	\centering
	\subfigure[Validation Loss]{
		\label{fig:resnet20-validation-loss}
		\includegraphics[width=0.48\textwidth]{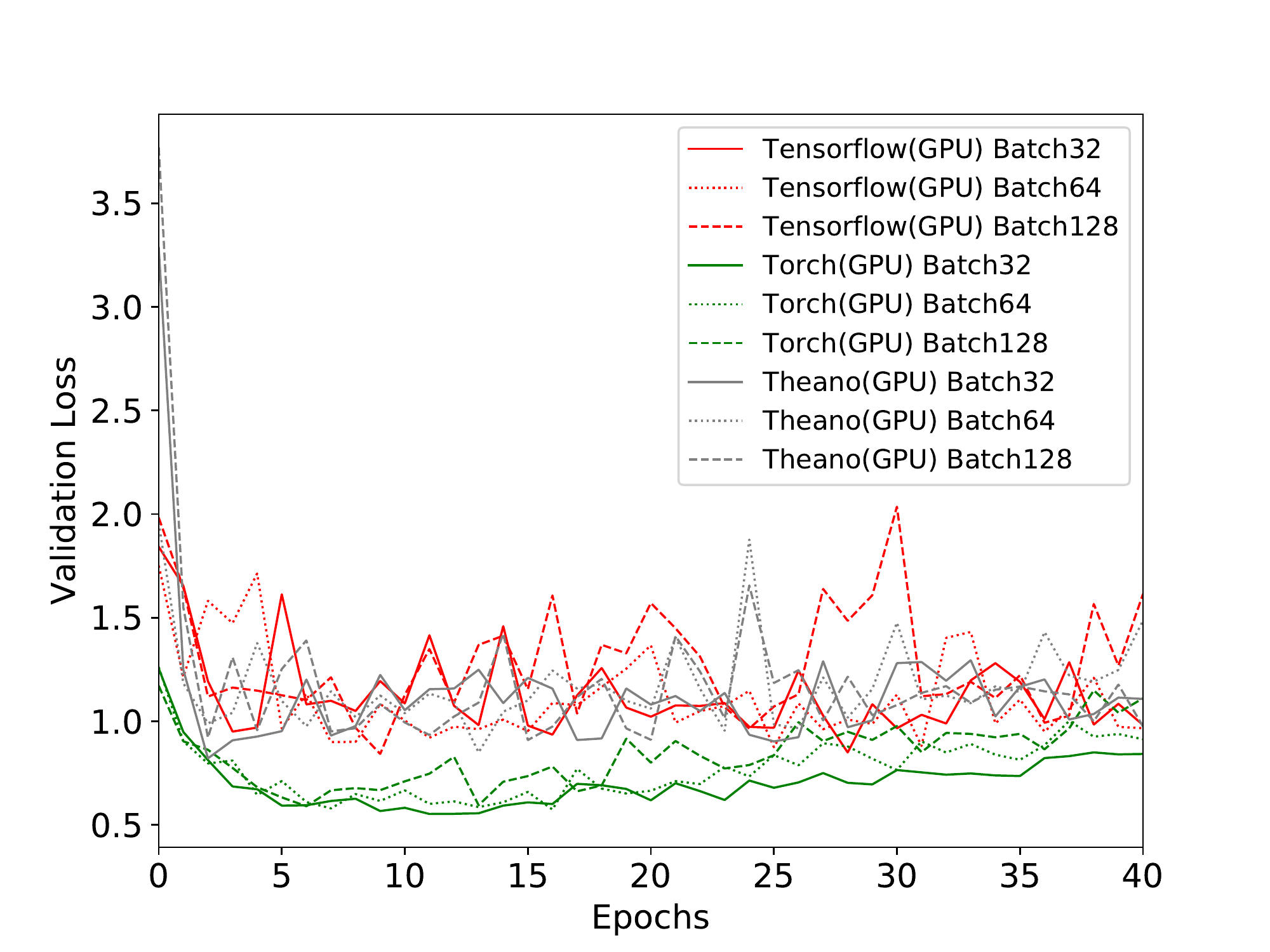}
	}
	\subfigure[Validation Accuracy]{
		\label{fig:resnet20-validation-acc}
		\includegraphics[width=0.48\textwidth]{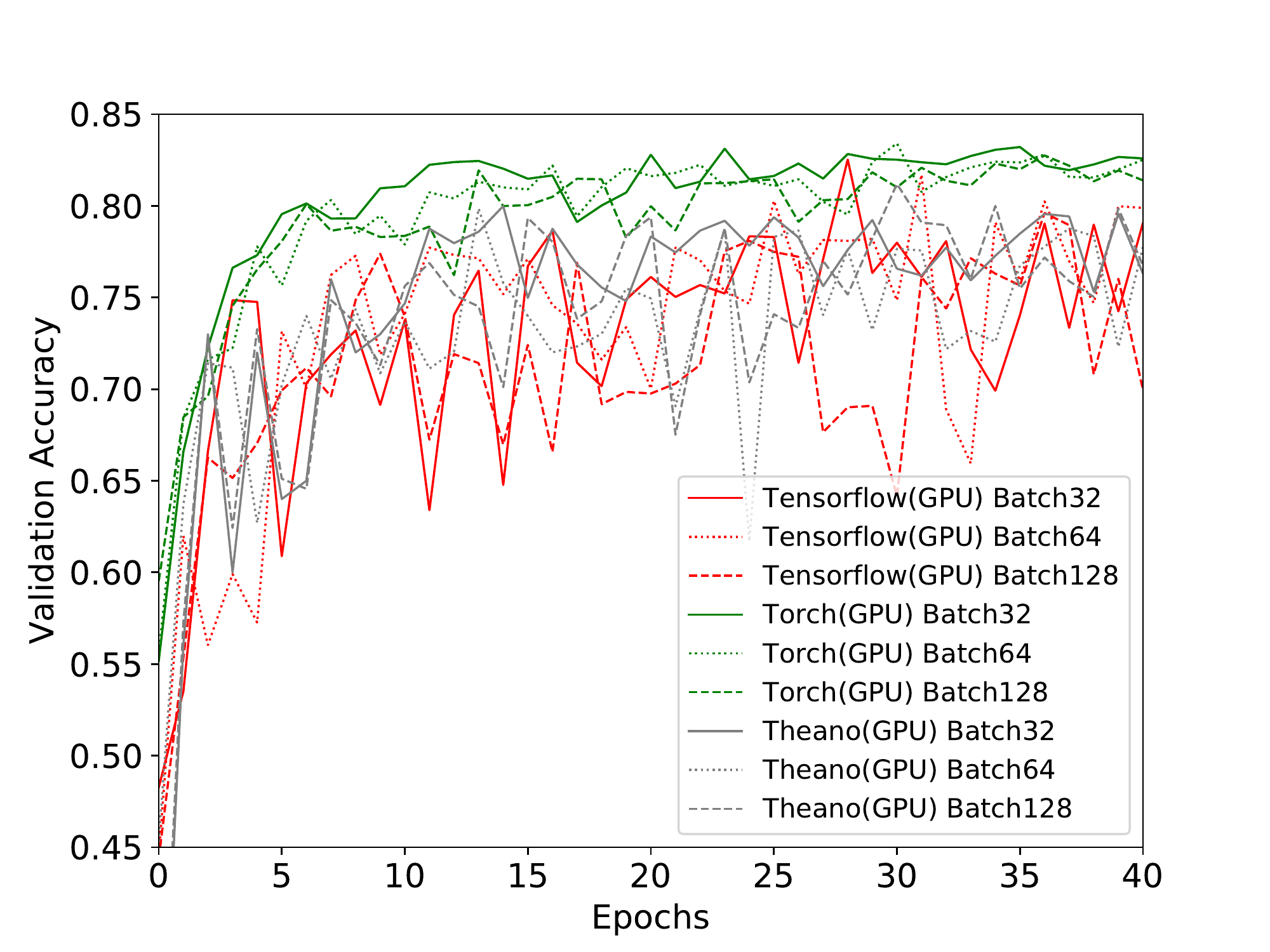}
	}
	\caption{Validation Loss and Accuracy of ResNet-20 with different training batch sizes and frameworks running on GPU}
	\label{fig:validatingloss}
\end{figure}

\noindent\textbf{Validation Loss and Accuracy}.
Figure~\ref{fig:validatingloss} shows the validation loss and accuracy for ResNet-20 with different configurations on GPU. Note that in order to make the comparative study more controllable on different frameworks, we only use the deterministic configuration in training process, and other advanced techniques such as data augmentation are not adopted. Hence, the validating accuracy of all CIFAR-10 models under different frameworks are a little lower. 
Overall, the validation plots of Torch are more stable than the other two frameworks whose plots have very large amplitudes. Meanwhile, the validation accuracy also reflects the same trend. As shown in Figure~\ref{fig:resnet20-validation-acc}, the Torch model converges after about 20 epochs while it is still changing on others even after 100 epochs. More importantly, Torch exhibits higher validation accuracy than TensorFlow and Theano at the same epoch. 

Comparing the training loss in Figure~\ref{fig:resnet20-trloss},  we find a larger training batch size leads to smaller training loss under Torch and TensorFlow on ResNet-20, which conversely bring in a higher validation loss, as depicted in Figure~\ref{fig:resnet20-validation-loss}. Situations on VGG-16 are similar. When it comes to LeNet-1 and LeNet-5, however, things are quite different. Torch shares the same characteristics with ResNet-20, but lower training and validation loss emerge with a larger training batch size, for both TensorFlow and Theano. The details of this investigation can refer to our website.

\noindent\textbf{Training Time and Model Size}.
Table~\ref{tab:traininigtime} shows the comparative results of training time and model size on GPU within different frameworks. Column {\it Model size} shows the average model size for each DNN. {\it Time Per Epoch} represents the average training time for each epoch under different training batch sizes (i.e., B-32, B-64 and B-128). The results show that, for each DNN, TensorFlow and Torch share similar model size while the Theano models are much bigger. For example, the model size of LeNet-1 for TensorFlow and Torch are 16KB and 14KB, respectively, while the size of Theano model is 65KB (about 4 times). For training time in each framework, the average time used for each epoch will decrease as the training batch size increases. Comparing the training time across different frameworks, we observe that Theano will be much slower than TensorFlow and Torch. Torch and TensorFlow can perform better than the other one under certain configurations.

\begin{tcolorbox}[size=title]
{\textbf{Answer to RQ1:} The runtime training behavior and performance are quite different under different DL frameworks. Generally, Torch outperforms TensorFlow and Theano during training stage, with smaller training loss, higher training accuracy and more training stability. And models of TensorFlow and Torch have smaller size than Theano. GPU acceleration for Theano is not well supported in our evaluated settings, consuming much time in training.} 
\end{tcolorbox}

\subsection{RQ2: Prediction Performance on Different Frameworks}
The pre-trained models will be deployed on different frameworks or platforms for predictive use. This section investigates the prediction performance on distinct DL frameworks and platforms. The predication accuracy can refer to the plots of validation accuracy in Section~\ref{sec:training-performance}. 
The study will be performed in following two aspects: 1) prediction performance deployed on different platforms. 
We use a server with GPU as the desktop application, a laptop as the web application platform, and a Huawei Nexus 6P, a Motorola Nexus 6, along with an IPhone 6S as the mobile applications. 2) prediction performance on different frameworks in the desktop application.

\begin{table}[]
\centering
\caption{Prediction time (GPU) with input batch size \protect 1000 and 10000 under different frameworks}
\label{tab:predict-metrics-across-frameworks}
\vspace{1ex}
\begin{tabular}{|c|c|c|c|c|c|c|}
\hline 
\multirow{2}{*}{Model} & \multicolumn{2}{c|}{TensorFlow(s)} & \multicolumn{2}{c|}{Theano(s)} & \multicolumn{2}{c|}{Torch(s)}\tabularnewline
\cline{2-7} 
 & 1000 & 10000 &1000 & 10000 & 1000 & 10000\tabularnewline
\hline 
\hline 
LeNet-1 & 0.088 & 0.05 & 0.358 & 0.402 & 0.007 & 0.010\tabularnewline
\hline 
LeNet-5 & 0.034 & 0.103 & 0.291 & 0.318 & 0.006 & 0.012\tabularnewline
\hline 
ResNet-20 & 0.306 & 1.229 & 9.139 & 18.169 & 0.044 & O/M\tabularnewline
\hline 
VGG-16 & 0.796 & 3.667 & 8.708 & 34.537 & 0.137 & O/M\tabularnewline
\hline 
\end{tabular}
\end{table}

Table~\ref{tab:predict-metrics-across-frameworks} shows the detailed prediction time of different models on GPU. Note that the prediction time only refers to time spent during the predicting operator, other procedures such as data loading and prepossessing are not included in the prediction time. For each dataset, we first randomly select 1000 images from the validation data to evaluate the performance with the input batch size 1000. Then we use the entire validation data all at once to evaluate the performance. Herein, $O/M$ represents the  result of {\it cuda out of memory} that is output by Torch when predicting with 10000 input batch size on CIFAR-10.

Results show that Torch predicts much faster than other two frameworks. Specifically, prediction under Torch is 5 to 10 times faster than TensorFlow and 26 to 60 times faster than Theano. As GPU support on Theano is not well, it is relatively much slower than other two frameworks.  

In addition, except for the study on models that are trained by different frameworks, we perform another study on models with the same weights. In details, after training under Theano with the wrapper Keras~\cite{keras}, we use the converter MMdnn~\cite{MMdnn} to transform the pre-trained model to Torch and TensorFlow.  A differential testing is conducted on these equal models. The target is to evaluate whether there are some errors (i.e., differences) in some frameworks or in the model converter. In this study, predictions of all converted models are the same on the test data. We also discover a potential bug within the converter MMdnn, when transforming from a Keras model to a Torch one. More details can be found on our website.     
  
\begin{tcolorbox}[size=title]
{ \textbf{Answer to RQ2:} 
Given DL models with the same design and runtime training configuration, the model prediction accuracy under different DL frameworks appear to be similar for certain epochs. However, the predictive time consumption are often quite different. Among the studied DL frameworks, Torch tends to be the most efficient in many cases while Theano is the slowest one (with approximately 26 to 60 times slower than Torch). When the state-of-the-art DL model conversion is used, the predication accuracy of the converted DL model remain to be the same across the three studied DL frameworks.} 
\end{tcolorbox}

\begin{figure*}[t]
    \centering
    \includegraphics[width=1.0\textwidth]{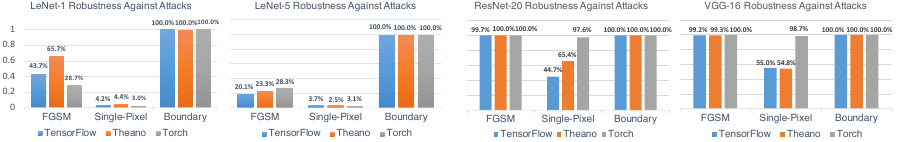}
    \caption{The robustness evaluation of DL models against adversarial attacks}
    \label{fig:robustness}
\end{figure*}

\begin{table}[!t]
\centering
\caption{Models selected for robustness measure}
\label{tab:robustnessmodels}
\begin{tabular}{|c|c|c|c|c|c|c|}
\hline 
\small
 \multirow{2}{*}{Model} & \multicolumn{2}{c|}{TensorFlow} & \multicolumn{2}{c|}{Theano} & \multicolumn{2}{c|}{Torch}\tabularnewline
 \cline{2-7}
 & \#epoch & Acc & \#epoch & Acc & \#epoch & Acc\tabularnewline
\hline 
\hline 
LeNet-1 & 50 & 98.5\% & 50 & 98.5\% & 50 & 98.6\%\tabularnewline
\hline 
LeNet-5 & 60 & 98.8\% & 60 & 99.0\% & 60 & 98.5\%\tabularnewline
\hline 
ResNet-20 & 70 & 83.1\% & 80 & 83.6\% & 85 & 83.9\%\tabularnewline
\hline 
VGG-16 & 85 & 82.6\% & 95  & 83.4\% & 95 & 86.4\%\tabularnewline
\hline 
\end{tabular}
\end{table}

\subsection{RQ3: DL Model Robustness}
This section investigates the robustness of models trained from different frameworks. For each dataset, the models trained on different frameworks are selected with two criterias: 1) following the common machine learning training practice~\cite{krizhevsky2012imagenet,goodfellow2016deep}, we select a relatively best trained model (i.e., higher accuracy without over-fitting and under-fitting) and 2) the training epochs of selected models are as close as possible to ensure the fairness. Table~\ref{tab:robustnessmodels} shows the details of models we select for comparing in robustness investigation.

For each selected model, we use the reliability against adversarial examples to measure its robustness. Specifically, we attack the model with the existing tool foolbox~\cite{rauber2017foolbox} and evaluate the success rate. Three kinds of representative attacks are attempted: FGSM~\cite{goodfellow6572explaining} (a gradient-based white-box attack), Single-Pixel-Attack~\cite{papernot2016practical} (a score-based black-box attack) and Boundary-Attack~\cite{brendel2017decision} (a decision-based black-box attack). Given an input, each attack generates test cases that result in an incorrect output from the DNN through the minor perturbations. The attacks are described as follows:
\begin{itemize}
    \item FGSM crafts an adversarial example $x' =x+\epsilon sign(g(x))$ by computing the gradient $g(x)=\nabla_X L(\theta,x,y)$, where $x$ is the given input and $y$ is the correct label of $x$. 
    \item Single-Pixel-Attack adds particularly-crafted noise to a single image pixel (e.g., changing a pixel from white to black), which no human would notice the difference to fool the DL network. 
    \item Boundary-Attack is one of the most effective decision-based adversarial attack to minimize the $L_{2}$\text{-}norm of adversarial perturbations. It does not reply on gradients or probabilities and finds the adversarial example with small perturbations.
\end{itemize}

We set $epsilons=100$ and $max\_epsion=0.1$ for FGSM, $max\_pixels=100$ for Single-Pixe-Attack, $iterations=100$ and $max\_directions=5$ for Boundary-Attack. Other parameters are set with the default values in foolbox. For each dataset, we randomly select 1000 images, which are predicted correctly by all the models, as the inputs of aforementioned attacks. To reduce randomness during the attack, each attack is repeated 10 times.

Figure~\ref{fig:robustness} shows the success rates of attacking models
trained from different frameworks, which represents the robustness accordingly. From the results, we can see Boundary attack achieves 100\% success rate in all models because it is the most effective decision-based adversarial attack~\cite{rauber2017foolbox}. 
This indicates that models trained from state-of-the-art frameworks are not robust enough to defend against the advanced attack~\cite{brendel2017decision}. Moreover, from the other two attacks, we can also compare the robustness to a certain degree. To compare the robustness roughly, we define the following equation to quantify robustness with the attacks.

\begin{equation*}
\small
\centering
\label{equa:probmutation}
\begin{array}{lr}
min = MIN(S_A^{m_1},\ldots,S_A^{m_n})\\
max = MAX(S_{A}^{m_1},\ldots,S_{A}^{m_n})\\
P(m_i,{A}) = 
\left\{
\begin{array}{lr}
(S_{A}^{m_i}-min)/(max-min) & if~min<max  \\
0 & if~min = max
\end{array}
\right.\\
R(m_i) = P(m_i,{A_1})+\ldots+P(m_i,{A_k}) \;\; k\geq 1
\end{array}
\end{equation*}
where $m_1,\ldots,m_n$ ($n\geq1$) represent the $n$ models for comparison, and ${A_1},\ldots,{A_k}$ are the $k$ types of attacks, $S_A^m$ represents the success rate of attack $A$ on the model $m$, and $R(m_i)$ quantifies the robustness of $m_i$ in terms of attacks $A_1,\ldots,A_n$, also known as the robustness indicator. The smaller value $R(m_i)$ is, the better robustness it exhibits. In this study, $m_1, m_2, m_3$ represent models from TensorFlow, Theano and Torch, respectively.

Calculated from the above equation, we find models from Torch and TensorFlow are relatively best for LeNet-1 and ResNet-20, respectively. Because they has the lowest attack success rates for FGSM and Single-Pixel-Attack. By contrast, the Theano model is relatively the worst one as it exhibits the highest attack success rates for LeNet-1. For LeNet-5, Theano model becomes best because $R(m_2)$ is the minimum (i.e., 0.39) while the Torch model is the worst with the $R(m_3)= 1.5$ being the maximum value. For VGG-16, TensorFlow is the best as $R(m_1)=0.005$ is the minimum while Torch model behaves the worst as $R(m_3)= 2$ is the maximum. Details about the robustness indicators can refer to our website.

\begin{tcolorbox}[size=title]
{ \textbf{Answer to RQ3:}  In summary, DL models that are trained from the studied frameworks are still vulnerable to adversarial attacks. In addition, the DL models with the same design and configuration trained from different frameworks exhibit different robustness. For different DL designs, the robustness are different as well.} 
\end{tcolorbox}

\subsection{RQ4: Prediction Performance on Different Platforms}
This section studies the prediction performance on different platforms and the effect of model quantization. For each DNN, we select one best-trained TensorFlow model which is then converted to a series of variants running on different platforms. Specifically, we use {\textit{TensorFlow Lite Converter}}, a converting tool officially supported by developer team, to shift TensorFlow graphs into TensorFlow Lite graphs. {\textit{TensorFlow.js Converter}} is used to load a pre-trained TensorFlow SavedModel into the browser and perform inference through TensorFlow.js. {\textit{Core ML Community Tools}} is a set of tools for Core ML model conversion and validation.

Table~\ref{tab:quantization} shows the detailed results of predication performance on popular platforms and the effect of quantization on mobile devices. Column {\it Platform} lists four platforms including PC, Web, Android and iOS. Column {\it Device} lists the frameworks or devices for each platform. And column {\it Quan.} represents whether the current model is quantized. Note that, the quantization in this study is only performed on mobile devices. Finally, columns {\it Size}, {\it Acc.} and {\it Pred.Time} show the size of each model and the prediction performance.
    
\subsubsection{Prediction on Different Platforms}\label{sec:accross-platforms}
For each DNN, we first compare the performance of each model before quantization on different platforms. Hence, only considering the rows in which column {\it Quan.} is {\it No}), we find that the model size does not change a lot after converting the TensorFlow model for web or mobile applications. Size of each web model will increase a little after the conversion (e.g., the model size increase from 16 KB to 20KB in LeNet-1). Meanwhile, the prediction accuracy on test data does not change after the conversion, which indicates current converters will not lose accuracy without quantization. 

\begin{table*}[htb]
\centering
\caption{Prediction performance on different platforms}
\label{tab:quantization}
\scriptsize
\setlength\tabcolsep{2.7pt}
\begin{tabular}{|c|c|c|c|c|c|c|}
\hline
Model & Platform & Device & Quan. & Size & Acc.(\%) & Pred.Time(s) \\ \hline\hline
\multirow{8}{*}{LeNet-1} & PC & TensorFlow & No & 16KB & 98.70 & 0.05 \\ \cline{2-7} 
 & Web & Chrome & No & 20KB & 98.70 & 16.03 \\ \cline{2-7} 
 & \multirow{4}{*}{Android} & \multirow{2}{*}{Nexus 6P} & No & 15KB & 98.70 & 4.19 \\ \cline{4-7} 
 &  &  & Yes & 5.4KB(\textbf{$-$64\%}) & 98.69 & 3.32 \\ \cline{3-7} 
 &  & \multirow{2}{*}{Nexus 6} & No & 15KB & 98.70 & 5.33 \\ \cline{4-7} 
 &  &  & Yes & 5.4KB(\textbf{$-$64\%}) & 98.69 & 3.80 \\ \cline{2-7} 
 & \multirow{2}{*}{iOS} & \multirow{2}{*}{Iphone 6S} & No & 16KB & 98.70 & 235.66 \\ \cline{4-7} 
 &  &  & Yes & 8KB(\textbf{$-$50\%}) & 98.70 & 238.27 \\ \hline\hline
\multirow{8}{*}{LeNet-5} & PC & TensorFlow & No & 178KB & 99.13 & 0.10 \\ \cline{2-7} 
 & Web & Chrome & No & 188KB & 99.13 & 21.22 \\ \cline{2-7} 
 & \multirow{4}{*}{Android} & \multirow{2}{*}{Nexus 6P} & No & 176KB & 99.13 & 6.16 \\ \cline{4-7} 
 &  &  & Yes & 50KB(\textbf{-71.2\%}) & 99.13 & 4.26 \\ \cline{3-7} 
 &  & \multirow{2}{*}{Nexus 6} & No & 176KB & 99.13 & 8.31 \\ \cline{4-7} 
 &  &  & Yes & 50KB(\textbf{-71.2\%}) & 99.13 & 5.30 \\ \cline{2-7} 
 & \multirow{2}{*}{iOS} & \multirow{2}{*}{Iphone 6S} & No & 180KB & 99.13 & 245.62 \\ \cline{4-7} 
 &  &  & Yes & 94KB(\textbf{-47.8\%}) & 99.13 & 248.92 \\ \hline\hline
\multirow{8}{*}{ResNet-20} & PC & TensorFlow & No & 1.1MB & 83.05 & 1.23 \\ \cline{2-7} 
 & Web & Chrome & No & 1.1MB & 83.05 & 201.22 \\ \cline{2-7} 
 & \multirow{4}{*}{Android} & \multirow{2}{*}{Nexus 6P} & No & 1.1MB & 83.05 & 495.21 \\ \cline{4-7} 
 &  &  & Yes & 290KB(\textbf{-74.3\%}) & 83.06 & 262.24 \\ \cline{3-7} 
 &  & \multirow{2}{*}{Nexus 6} & No & 1.1MB & 83.05 & 565.30 \\ \cline{4-7} 
 &  &  & Yes & 290KB(\textbf{-74.3\%}) & 83.06 & 320.41 \\ \cline{2-7} 
 & \multirow{2}{*}{iOS} & \multirow{2}{*}{Iphone 6S} & No & 1.1MB & 83.09 & 374.73 \\ \cline{4-7} 
 &  &  & Yes & 557KB(\textbf{-50.6\%}) & 83.05 & 383.49 \\ \hline\hline
\multirow{8}{*}{VGG-16} & PC & TensorFlow & No & 129MB & 84.20 & 3.67 \\ \cline{2-7} 
 & Web & Chrome & No & 134.8MB & 84.20 & 453.58 \\ \cline{2-7} 
 & \multirow{4}{*}{Android} & \multirow{2}{*}{Nexus 6P} & No & 129MB & 84.21 & 2201.95 \\ \cline{4-7} 
 &  &  & Yes & 33MB(\textbf{-74.4\%}) & 84.20 & 1996.54 \\ \cline{3-7} 
 &  & \multirow{2}{*}{Nexus 6} & No & 129MB & 84.20 & 2432.51 \\ \cline{4-7} 
 &  &  & Yes & 33MB(\textbf{-74.4\%}) & 84.20 & 2055.34 \\ \cline{2-7} 
 & \multirow{2}{*}{iOS} & \multirow{2}{*}{Iphone 6S} & No & 137.8MB & 84.19 & 1699.90 \\ \cline{4-7} 
 &  &  & Yes & 67.4MB(\textbf{-51.1\%}) & 84.22 & 1768.87 \\ \hline
\end{tabular}%
\end{table*}

Prediction time of models across platforms are totally different. For LeNet-1 and LeNet-5, predicting with Android devices are much faster than that on web side (about 4 times) and iOS device (about 50 times). For example, predicting 10000 test data will spend 16.03 seconds on chrome, 235.66 seconds on Iphone 6S but only 4.19 seconds on Nexus 6P. Surprised by such a big difference among platforms, we carefully recheck the results and confirm its correctness. For ResNet-20 and VGG-16, the results seem easy to understand. Prediction on PC with GPU is very fast. Then the web application is faster than mobile devices as it is also deployed with computer servers as backend support. Different from LeNet-1 and LeNet-5, prediction on iOS is faster than Android when it comes to ResNet-20 and VGG-16. This seems to indicate that as the complexity of the model increases, the performance advantage of iPhone gradually begins to appear.

\subsubsection{Prediction with Quantization}
Quantization is a technique to optimize the model so that it can be run on the mobile devices more faster~\cite{odena2018tensorfuzz, 2018arXiv180901266X}. Considering the rows in which column {\it Quan.} is {\it Yes}), we find that the model size will decrease 50\% to 74.4\% after quantization. It will save much storage and memory for mobile devices. It can be obviously seen from the results, quantization almost does not affect the predication accuracy. Specifically, the accuracy of quantized model will reduce up to 0.01\% on Android device and 0.04\% on iOS device. Even in some cases, the accuracy of quantized model will be higher, e.g., the quantized ResNet-20 model on Android devices increases 0.01\% while the quantized VGG-16 model on iOS device rises 0.03\%. Predicting on Android devices after quantization is faster than the previous model, the improvement is more obvious for larger models (e.g., ResNet-20 and VGG-16). However, quantization on iOS will slow down the prediction speed a little, revealing an interesting phenomenon for further exploration and improvement.

We also try to evaluate the energy consumption during runtime prediction on mobile devices. However, it is difficult to scratch accurate battery losses for ordinary developers from mobile systems. Instead, we record the battery changes from screen. Specifically, we perform 5 replicated evaluations for each model (both the original model and the quantized one), and record the total power loss for further comparison. It is necessary to note that, the battery consumption after 5 experiments is not so apparent as expected for either LeNet-1 or LeNet-5, no matter quantized or not. This may be because the models of LeNet-1 and LeNet-5 are  relatively lightweight, making no so much complex computation at runtime. As with VGG-16, validation 10000 CIFAR-10 images on the original model, we get an approximately 15\% battery decline on Nexus 6P, 16\% on Nexus 6 and 18\% on Iphone 6S. After quantization, the power loss on Nexus 6P, Nexus 6 and Iphone 6S are 14\%, 15\% and 19\%, respectively. Cases for ResNet-20 are quite similar, almost exhibiting no difference with VGG-16. As can be seen, model quantization on Android devices indeed reduces the computing overheads by small degrees, while the situation on Iphone 6S is just opposite, which is worth further investigation. Although roughly estimated, the results still provide us some useful feedbacks about the energy consumption before and after quantization.

\begin{tcolorbox}[size=title]
{ \textbf{Answer to RQ4:} The prediction accuracy of the same model under different DL platforms are nearly the same but the prediction time efficiency is quite different. In some cases, the prediction on mobile device can be much faster than web application executed on a normal PC. The existing quantization technique is effective to reduce the model size while preserving the accuracy in most cases. The prediction time after quantization is much shorter in Android devices while it takes longer for iOS device. Situation on power consumption is similar. The platform specific DL execution acceleration optimization needs to be carefully considered.}
\end{tcolorbox}
\vspace{-3mm}

\subsection{Discussion}
We perform a large scale comparative empirical study on DL frameworks and platforms. Through this study, we found: 1) given the same DNN architecture and runtime training configuration, differences are shown across frameworks in terms of runtime training performance, DL model robustness, as well as prediction performance. 2) given the same model using state-of-the-art conversion techniques, although the prediction accuracy tends to be preserve, the runtime efficiency shows to be quite different.

Although many DL frameworks and platforms exist, the current DL software development still lacks systematic engineering guidance. The DL models are still vulnerable to adversarial attacks, which is confirmed in our evaluation in all DL frameworks.  And the standards, benchmarks and testing techniques for DL software development are still immature and needed to be further investigated. Besides, robust DL model conversion techniques (e.g., MMdnn, ONNX) are also urgently needed in academia and Industry. It would be a big challenge to provide reliable guarantee for DL model compatibility and equivalence conversion. Further debugging and testing technique and toolchain support are also in demand.

With the large demand for intelligent application on mobile devices, the platform migration and customization become a must. Quantization is a technique to preserve the prediction accuracy with lower overheads. Current quantization techniques reduce the model size and prediction time, but the performance is still far from satisfactory in faces of complex models, which requires further enhancement.
\section{Related Work}
In this section, we review the previous work in following three aspects: study on traditional software platforms, study on deep learning frameworks, and study on DL platforms.

\subsection{Study on Traditional Software Platforms}
S.Z.S.Idrus et al.\cite{idrus2008performance} compared on how an encryption algorithm (exclusive OR) is implemented in different browsers (IE, Firefox, Opera and Netscape Navigator) and determined which algorithm works most compatibly with which web browser.
A.Charland et al.\cite{charland2011mobile} tested the performance of JavaScript on different mobile platforms (iOS 4 and Android 2.2) by using two JavaScript benchmarks SunSpider and V8, respectively.
S.Narayan et al.\cite{narayan2009performance} reported the difference of IPv4 and IPv6 network performance between Windows Vista and Ubuntu. By evaluating throughput, delay, jitter and CPU usage, they discovered that platform has implications on network performance.
A.Algirda et al.\cite{avizienis1988search} utilized 6 mainstream programming languages, and implemented a fault-tolerant flight control software. Based on this, they counted the code lines, modules, statements and the mean number of statements per modules, and analyzed the differences among these languages.
L. Prechelt \cite{prechelt2000empirical} focused on the phone-code program and compared the performance between C, C++, Java, Perl, Python, Rexx and Tcl, in terms of the program length, programming effort, runtime efficiency, memory consumption, and reliability.
S.Nanz et al.\cite{nanz2013benchmarking} compared four markedly different approaches to parallel programming (i.e., Chapel, Cilk, Go and Threading Building  Blocks).  Each  language  is  used  to implement sequential and parallel versions of six benchmark programs.

\subsection{Study on Deep Learning Frameworks}
The rapid emergence of deep learning frameworks attracts researchers' attention on the performance of frameworks.
S.Bahrampour et al.\cite{bahrampour2016comparative} presented a comparative study on five deep learning frameworks (i.e., Caffe, Neaon, TensorFlow, Theano and Torch) and evaluated the forward time and gradient computation time to asses the framework performance. 
P.Druzhkov et al.\cite{druzhkov2016survey} made a survey of deep learning software tools for image classification and object detection. They compared up to 15 kinds of deep learning tools for these specific tasks.
S.Shams et al. \cite{shams2017evaluation} analyzed Caffe, TensorFlow and Apache SINGA over several hardware environments. In order to investigate the performance, they measured the time per training iteration and the number of images trained with in a millisecond for comparison.
K.Kochura et al.\cite{kochura2017comparative} compared the basic features (i.e., GPU
support, GUI, operating systems and language support) of TensorFlow, Deep Learning4j and H2O and conducted throughout performance tests. In particular, H20 was tested under single threaded mode and multi-threaded mode.
D.Li et al. \cite{li2016evaluating} evaluated the energy efficiency of CNNs on CPUs and GPUs by calculating the energy and power consumption of 10 frameworks (K20-Torch, Tx-Caffe et al.).
S.Shaohuai et al. \cite{shi2016benchmarking} calculated the time per mini-batch with different threads (i.e., 1, 2, 4, 8) and models (FCN-S, ResNet-50 et al.) within Caffe, CNTK, TensorFlow, MXNet and Torch. Additionally, benchmarks of these frameworks are proposed.

Compared to these works, our empirical study conducts a more comprehensive analysis including training performance, prediction performance, robustness and model optimization on different frameworks.

\subsection{Study of Deep Learning Platforms}
There is a tendency of transplanting deep learning frameworks onto heterogeneous platforms for a broader application.
O.Kaoru et al. made a survey \cite{ota2017deep} on deep learning for mobile multimedia use and introduced the the low-complexity DL algorithms, an optimized software framework for mobile environments and the specialized hardware for supporting the computationally expensive
processes of deep network training and inference. By contrast, the performance of models after quantization is considered in our study as an evaluation metric which is insufficient in O.Kaoru's work.
AI-Benchmark\cite{AIBenchmark} proposed a AI performance ranking for current mainstream mobile phones. Nine testing tasks such as object recognition and face recognition are used as criteria for performance judging.
O.Alsing et al. \cite{alsing2018mobile} summarized the latest mobile object detection methods using TensorFlow Lite and analyzed the performance and latency payoff of different deep learning models on mobile devices.
And J.Wang\cite{wang2018deep} et al. provided an overview of the current achievements about mobile deep learning technology and applications. 
In this work, we not only conduct evaluations on the mobile devices, as demonstrated in previous work, but also shift the testing on the web browsers. In other words, we have more widely investigated the differences on various deep learning platforms.
\section{Conclusion and Future Work}
Deep learning has become one of the leading technology for future intelligent software solutions.  And DL frameworks act as the driving wheels for constructing deep learning softwares. Although many DL frameworks are currently available for academic and industry, their difference and incompatibility in computational paradigm, architecture design, and implementation bring new challenges for the DL based software production process, including development, deployment, maintenance, migration, and software reuse, etc. 

In this paper, we initiate the first step to investigate how existing DL frameworks and platforms influence the development and deployment of DL softwares in multiple perspectives. We find that it has turned out to be a pressing concern and urgent pain point for the application across heterogeneous frameworks and platforms. The exact same network design and training configuration may often result in different training and prediction performance, and robustness issues. The incompatibility error and conversion loss would arise when migrating a DL model from one framework to another. In addition, the universal DL solutions across platforms are desperately on demand, especially for mobile use. Our work makes the first step along this direction towards building universal DL software across various platforms. 
One of our main future work is to propose compatible solutions for DL framework implementation inconsistencies. We also plan to make more investigations on how static and dynamic computational paradigm impact the DL development process. We hope our work draws the attention of DL software engineering community, altogether to address the urgent demands towards future intelligent of things for everyone.

\bibliographystyle{IEEEtran}
\bibliography{ref}
\end{sloppypar}
\end{document}